\documentclass[runningheads]{llncs}
\usepackage{graphicx}
\usepackage{multirow}
\usepackage{color}
\usepackage{subcaption}
\begin{document}
\title{ClusterTabNet: Supervised clustering method for table detection and table structure recognition \\ \mbox{ }}

\titlerunning{ClusterTabNet}
%
\author{Marek Polewczyk \inst{1} \and
        Marco Spinaci \inst{1}}

\authorrunning{M. Polewczyk and M. Spinaci}

\institute{SAP Business AI
\email{\{marek.polewczyk,marco.spinaci\}@sap.com}}

\maketitle              
\begin{abstract}
Table detection and recognition consists of locating tables within a given document and identifying the exact location of its pieces, such as rows, columns, and headers. We present a novel deep-learning-based method\footnote{The code is released at \url{https://github.com/SAP-samples/clustertabnet}} to cluster words in documents which we apply to detect and recognize tables given the OCR output. We interpret table structure bottom-up as a graph of relations between pairs of words (belonging to the same row, column, header, as well as to the same table) and use a transformer encoder model to predict its adjacency matrix. We demonstrate the performance of our method on the PubTables-1M dataset introduced in \cite{pubtable1m} as well as PubTabNet and FinTabNet datasets. Compared to the current state-of-the-art detection methods such as DETR \cite{detr} and Faster R-CNN \cite{faster_rcnn}, our method achieves similar or better accuracy, while requiring a significantly smaller model.

\end{abstract}
\section{Introduction}
\label{sec:introduction}

The problem of detecting tables in documents and recognizing their structure is a long-standing application of machine learning techniques, as it is a necessary step towards automatically understanding the contents of a document. Early methods for example \cite{heuristic_distance,heuristic_spacing} involved heuristic-based approaches to detect tables within a page and recognize its contents; but due to the variety of styles of both tables and of the documents they appear in, these approaches cannot generalize arbitrarily. Therefore, recently a new wave of deep learning-based approaches has emerged.

Most of these approaches take the document image as input and directly deal with table detection and recognition as an object detection task. For instance, they apply off-the-shelf state-of-the-art algorithms such as DETR \cite{detr} or Faster R-CNN \cite{faster_rcnn} to the raw image. This approach, however, has several disadvantages: first of all, the models tend to be large, and their application is expensive since a high resolution is often required. Secondly, in order to rely on hints, such as the word ``Table'' in a caption, it basically has to re-learn the OCR task. Finally, object detection methods rely on bounding boxes, which fail for pages that are not perfectly aligned (e.g. rotated scans, phone pictures).

More recently, other approaches in the wake of LayoutLM \cite{xu2019layoutlm} circumvent the need to have a model re-inventing OCR by implementing a multi-modal transformer consuming both image patches and post-OCR words with bounding boxes. While this solves the second problem above, in the original application it was limited to various classification problems, and not clustering tasks, such as table recognition.

We propose an alternative, novel way to detect tables and recognize their structure given the OCR output. Once words and their corresponding contents are extracted from the document, we use a transformer encoder neural network to cluster words into tables and table structure elements. To do so, we see the problems of both table detection and table recognition as a clustering task: given a page of text, words belonging to the same table are clustered together, while words not in tables are considered as not clustered. Recognizing the content of a table, that is, partitioning its contents into cells, rows, columns, and headers can be viewed in the same framework.

However, traditional algorithms for clustering cannot be used for this task, as the concept of belonging to the same cluster (e.g. the same table row) must be inferred from training on a similar dataset. Otherwise said, this is a supervised learning task. Similarly, one cannot hope to apply classification algorithms (to classify, say, words that are in the ``first'' table, ``second'' table, etc.), because there is no intrinsic feature that distinguishes any table from another in the same page or any row within a table. We are therefore looking at the topic of ``supervised clustering'', which is not often encountered in the literature. In this framework, each training point of the dataset consists of a multitude of individuals, together with information on which of these individuals are clustered together. For an analogous situation in computer vision, one can think of the task of image segmentation: each data point (an image) is a collection of pixels, together with information of which two pixels belong to the same object.

In this work, we propose to solve this task by seeing clusters as a graph and predicting the adjacency matrix directly. That is, if there are $n$ words in a page, we are trying to predict a graph with $n$ vertices, where each cluster is a connected component. In the simplest case where each word belongs to exactly one cell, one row, one column, etc., this is an undirected graph where each connected component is, in fact, a clique. To that aim, we use a transformer encoder architecture. The input consists of the set of all the word contents and bounding boxes coming from OCR. For each clustering target (tables, rows, columns, headers), our model outputs an adjacency matrix with size equal to the number of words in the page.

Experiments are conducted on the PubTable-1M \cite{pubtable1m}, ICDAR-2019 \cite{icdar2019}, FinTabNet \cite{global_table_extractor}, and PubTabNet \cite{image_based_table_recognition} datasets showing competitive, or superior, performance to state-of-the-art models. Several trade-offs between quality and speed are discussed, such as splitting the task into two sequential steps (detection + recognition) and adding image crops information to the input.

\section{Related Work}
\label{sec:relatedwork}

\subsection{Detection methods}

Table detection has been studied for decades and researchers invented different heuristic-based, machine learning-based, and deep learning-based algorithms.

Heuristic-based methods use various features such as lines, keywords, spacing, and alignment. For instance method proposed in \cite{heuristic_distance} considers the distance between the consecutive words to detect tables and in \cite{heuristic_spacing} authors presented a method that uses local thresholds for word spacing and line height for detecting table regions. Such rule-based approaches work well for documents with uniform layouts, however, they are not applicable to a wide variety of tables.

Machine learning-based algorithms are the next category of algorithms. They include unsupervised clustering of words \cite{unsupervised_clustering}, tree-based algorithms \cite{tree_bases}, decision tree, SVM classifiers \cite{decision_tree_svm}, etc.
Deep learning-based algorithms are currently the best method for detecting and recognizing tables as the approach works with good precision on a variety of document layouts. One of the most common approaches is to use object detection deep learning models that are based on CNN or vision transformers. In \cite{faster_rcnn_based_model,deepdesrt} authors have shown how to recognize tables with the Faster R-CNN-based model. In \cite{cascadetabnet} the CascadeTabNet model and in \cite{tablenet} the TableNet model were presented that detect and recognize tables and their structure using the end-to-end deep learning model. In \cite{yolo}, a table-detecting algorithm based on the YOLO principle is described. Many other models have been used for table detection and recognition over the years. Recently transformer-based models were applied to document layout analysis in \cite {pubtable1m,transformer_detection}.

Other approaches are based on image-to-text \cite{tablebank} or graph neural networks \cite{complicated_table_structure_recognition,rethinking_table_recognition_using_graph_neural_networks}. These techniques often include some custom rules and processing steps to perform table structure recognition.

Our solution is closer to the LayoutLM network \cite{xu2019layoutlm}, which takes word boxes as input; however, we used a novel clustering technique on top of that model.

\subsection{How we are different}
Our method relies on the OCR output. Since OCR models are already optimized to extract words positions and contents even in noisy documents, our model is automatically robust to printing or scanning defects. We do not require a deep vision model, nor heavy image data augmentation. Our model can concentrate on learning dependencies between word boxes that belong to a given table, row, column, or header. The problem then is no longer a computer vision task but rather a task to find relations between words, such as spacing. Our solution also allows for using image patches to teach the model to take into consideration features like table borders, which is useful for optimal results. Furthermore, the formulation is totally independent of the page orientation and does not require, for example, rows to be perfectly horizontal.

Moreover, our solution does not require a big model. We have used a 4-layer transformer encoder that is smaller $\sim$5M parameters (excluding embedding layers) versus  $\sim$29M parameters in DETR network. Despite the simple setup, our method can outperform other approaches in terms of accuracy.

\section{Transformers for supervised clustering}
\begin{figure}
    \centering
    \includegraphics[width=1.0\textwidth]{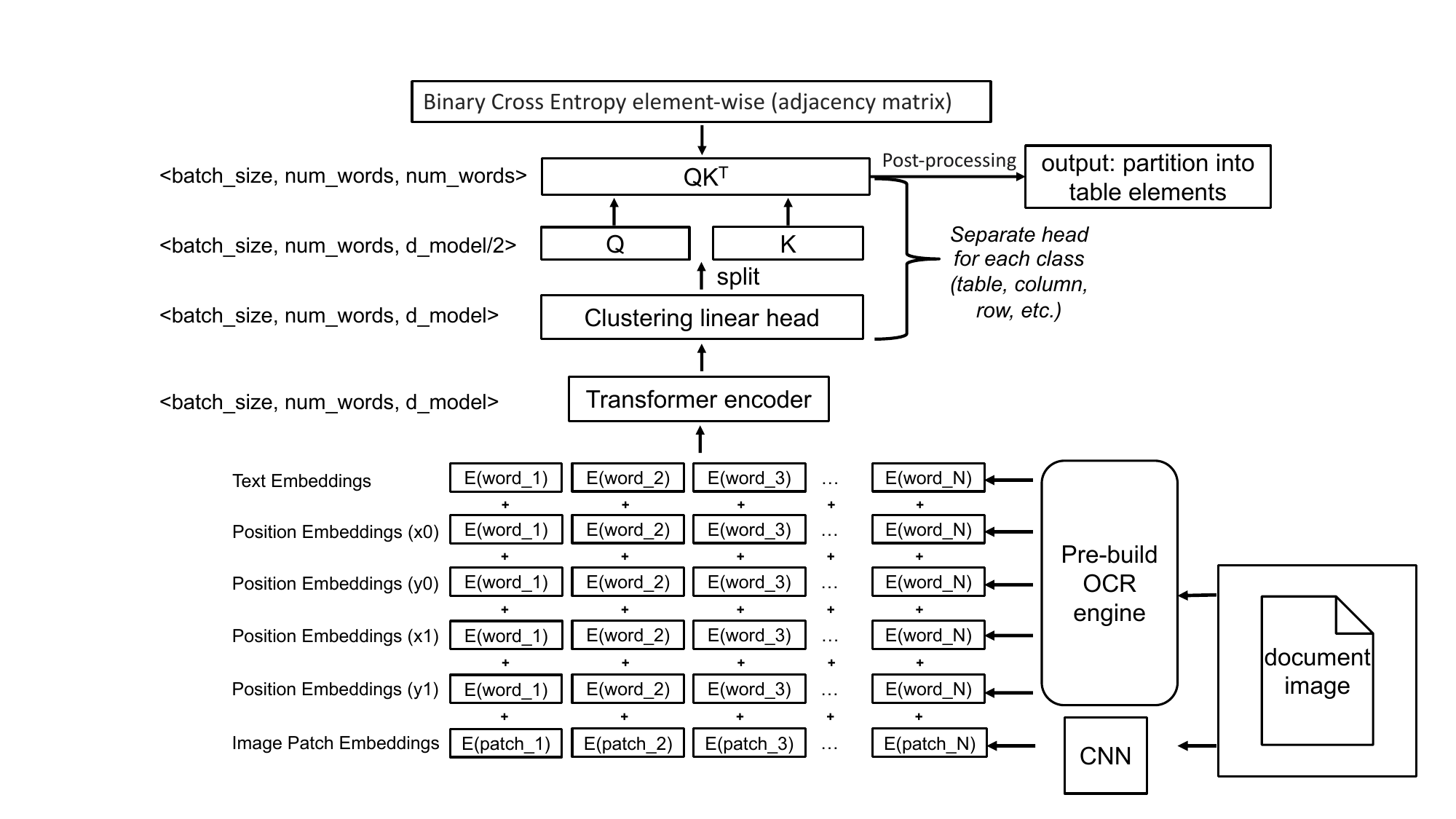}
    \caption{Schematic representation of the clustering network architecture with its input and output, as well as the loss applied during training.}
    \label{fig:architecture}
\end{figure}

Our model uses the standard transformer encoder as architecture; the only two differences concern positional encodings and the task-specific heads we describe below and visualize in Figure \ref{fig:architecture}. We use supervised learning to predict the adjacency matrix which determines how input words are clustered.

\subsection{Positional encoding and vocabulary embedding}
\label{subsec:pos_enc_voc_emb}
The input of the models are words, as output from an OCR engine. An OCR word is, typically, the combination of its content and a bounding box, itself represented as 4 numbers (the coordinates of the upper-left and lower-right corners), or more, if the OCR engine supports rotated words (either 5, in the form of center coordinates, plus width, height, and rotation in the $[0, 180)$ range; or 8, in the form of all vertex coordinates). In either case, we convert the coordinates (except rotation) to integers, by expressing them as percentages of the page width/height, multiplying by 1024 and rounding down to an integer in $[0, 1023]$. Similarly, we construct a dictionary of the most common words in the training set (we set the dictionary size to 30015, plus UNK). Each of the resulting integers (of which there are between 5 and 9) is embedded with a separate learnable embedding layer of the transformer hidden dimension, and the results are summed together and fed into the transformer encoder.

We also experimented with adding image crops to the OCR words. To do this, we crop out a padded bounding box around the word coordinates and rescale it to a fixed $32\times 32$ size. We pass it through a small ResNet-like CNN with 4 blocks, including two max pooling layers after the first two blocks. The output number of channels of each block is 16, 32, 48, and 80. The output tensor, of shape $(80, 8, 8)$, is then flattened and passed through a linear layer to reduce it to a vector of size $d_{model}$ (the transformer hidden dimension, which in our experiments defaults to 256). This is added to the outputs of the content and coordinates embeddings.

We also remark that, in NLP tasks, recently more success has been obtained by using subword level tokenizers, which has the advantage of avoiding out-of-dictionary issues. In our situation, however, we opted for full word, at the cost of adding a generic UNK term. We argue that, for the sake of table detection, it might be useful to detect common words (such as ``Table'' or common headers), but much less so for uncommon terms. Since switching from full words to subwords significantly increases the sequence length, we chose not to trade the speed and memory advantage for a minor potential gain.

We want to minimize the occurrence of UNK tokens especially when dealing with text in various languages, so we apply several normalization techniques to each word: represent all characters in ASCII characters, remove white spaces, replace all lowercase characters by 'a' and upper-case ones by 'A', all digits by '1' and other non-digit characters with ','. In this way, the final dictionary contains words out of only 4 possible characters. At the same time, we keep the information about uppercase characters which is useful for headers and numbers which are often stored as values in the table. For table detection and recognition, there should be no difference in what digits exactly are in the table, so such normalization is likely beneficial. In this way, we can avoid training in multiple languages (for example, via automated translations of the, mostly English, training sets).

\subsection{Output heads}

The output of the transformer encoder is fed into multiple, identical heads: one to group words into tables, one into rows, etc. In terms of architecture, each such head is the sequence of one linear layer, followed by a GELU activation, layer normalization, dropout, and another final linear layer. This is the same architecture as one could use, for example, for named entity recognition. The output of the last layer thus has shape $(N, L, D)$, where $N$ is the batch size, $L$ is the sequence length (i.e. number of words), and $D$ is some chosen output dimension.

The difference is that we split the output of the last layer into two identically sized tensors of shape $(N, L, D/2)$ (equivalently, one could of course use two identical linear layers after the dropout). We call these two tensors $K$ and $Q$ because they play a role similar to key and query in standard attention. The final output is $\sigma(QK^T)$, where $\sigma$ is the sigmoid function:
$$
\sigma(x) = \frac{1}{1 + e^{-x}}.
$$

\subsection{Training targets}
\label{sec:training-data}
We convert the ground truth data to an adjacency matrix, with the position $(i,j)$ being equal to 1 if and only if the two words are in the same cluster. This gives a total of five matrices (one for detection, plus four for recognition: cells, headers, rows, and columns). The relation of being part of the same table or the same cell is symmetric; therefore, the resulting matrix is also symmetric (i.e. the above equals also the value in position $(j, i)$). For headers, rows, and columns, however, we include the possibility of non-symmetric matrices, to support cells spanning multiple rows and/or multiple columns. If, say, cell ``A'' spans across rows ``B'' and ``C'', then every word belonging to cells entirely within row ``B'' (resp. ``C'') will have symmetric connections with every other word belonging to row ``B'' (resp. ``C''), and additionally a one-directional connection to words within cell ``A''. On the other hand, words within cell ``A'' will only point to other words in the same cell, also in the row head. The resulting graph is, therefore, directed.

\subsection{Training Loss}
The loss is computed by comparing the predicted values against adjacency matrix labels using elementwise binary cross-entropy: if the matrix at position $(i,j)$ has a predicted value after the sigmoid function of $p \in (0,1)$, then the loss is $-\log(p)$ if the label in that position is equal to 1, otherwise, the loss is $-\log(1-p)$ if the label in that position is equal to 0. The loss is computed elementwise on the square matrices (the predicted $\sigma(QK^T)$ and the adjacency matrix label), and then aggregated via sum after masking the padded elements if the document had fewer words than the fixed training sequence length.

\subsection{Post-processing}
\begin{figure}
    \centering
    \includegraphics[width=0.99\textwidth]{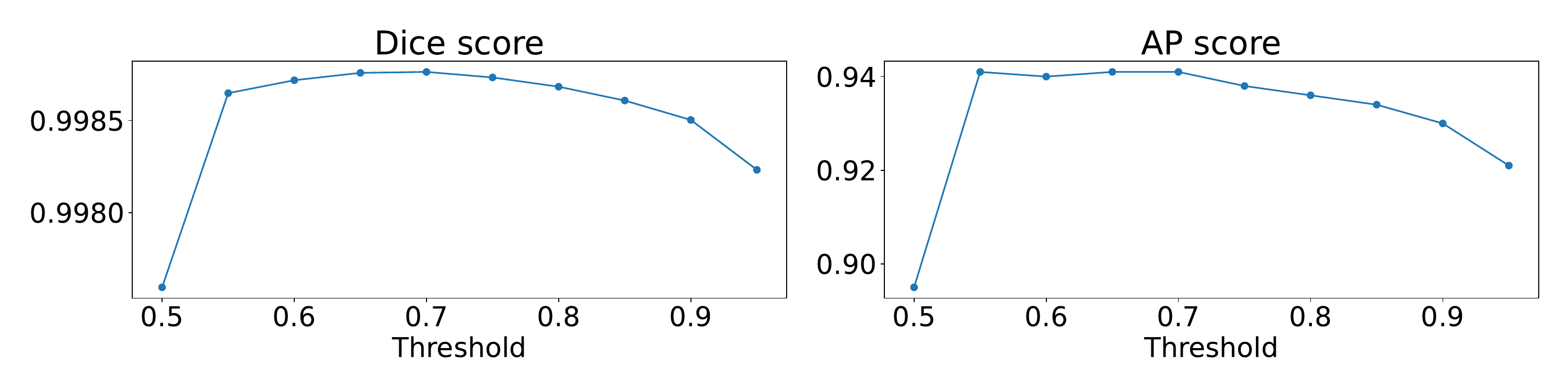}
    \caption{Dice score and average precision for various choices of the hard threshold.}
    \label{fig:threshold}
\end{figure}

During inference, we first find ``strong connections'' in the graph. That is, we average the prediction matrix with its transpose and threshold the results:
$$
\frac{1}{2}\big(\sigma(QK^T )+\sigma(KQ^T)\big)\geq k.
$$
We choose the fixed threshold value $k$ via validation (see Figure \ref{fig:threshold}).\footnote{The usual choice $k=0.5$ is actually not optimal: because of the connected components algorithm that follows thresholding, a single wrong ``1'' prediction ends up wrongly merging two clusters (e.g. two cells, or two rows), while a single ``0'' prediction is most likely fixed by the connected component algorithm. More robust, iterative approaches based on majority voting instead of relying on single connection could be used in alternative to connected components. However, preliminary tests suggested that such approaches would be considerably slower while bringing only marginal improvements, so we do not include them here.} The model may make mistakes, typically missing some 1s in the output (since labels are sparse, with a majority of 0s); we run a connected components algorithm on the result to fix the missing connections. ``Being in the same cluster'' should be a transitive relation: if the model predicted that $A$ should be clustered with $B$ and $B$ should be clustered with $C$, but the information that $A$ should be clustered with $C$ is not predicted correctly, then such mistake is fixed. Each connected component will be interpreted, depending on the head, as a table, a cell, a row, a column, or a header.

For the fields admitting spanning cells (rows, columns, and headers), we then proceed to look for weak (i.e. one-directional) connections. We first set values to 0 in the original, non-symmetrized, adjacency matrix in places where a strong connection has already been found. Then, we consider entries that are above the threshold of 0.5 as weak connections.\footnote{Spanning cells are too rare to run proper validation on this hyperparameter, so we accept the default choice.} To avoid having too many weak connections we consider a word to be a spanning cell only if more than half of the words in a strongly connected cluster (row, column, or header) have a weak connection pointing to that word in the spanning cell.


Finally, to reconstruct the bounding box coordinates of the table and of each component, we can simply take the maximum and minimum $x$ and $y$ coordinates of each word belonging to the corresponding cluster built by strong connections. If rotated pages are to be supported, standard algorithms exist to compute the minimum (rotated) bounding rectangle of a given set of points, which can then be used instead. Weak connections are used to extend rows only horizontally and columns only vertically, otherwise after extension columns and rows could overlap with each other and wrongly predicted weak connections would alter the result significantly.

\subsection{Sequence length}

The maximum sequence length can be chosen arbitrarily, based on the type of document to which one wants to apply this method, although too large values heavily affect runtime and memory usage, due to the $O(n^2)$ nature of attention. In practice, a common solution is to limit the number of words to some fixed value during training (cutting away redundant words or padding if fewer). At prediction time, if faced with a document that exceeds that amount, the document can be split into overlapping sets of words, each processed separately. Clusters can be computed separately and joined together (again using the transitivity principle above, even for words belonging to different page splits). In our experiments, we have fixed the maximum amount to 1000 words.

\section{Data}
\label{sec:data}

Recently, large datasets like SciTSR, TableBank, PubTabNet, and FinTabNet \cite{tablebank,global_table_extractor,image_based_table_recognition,complicated_table_structure_recognition} have been created by collecting crowd-sourced table annotations automatically from existing documents. Each dataset has annotations of table structures in HTML or XML format. Moreover, the SynthTabNet dataset \cite{synthtabnet} with synthetically generated table layouts and annotations has been generated.

We have used the PubTables-1M \cite{pubtable1m}, FinTabNet \cite{global_table_extractor}, PubTabNet \cite{image_based_table_recognition}, and SynthTabNet \cite{synthtabnet} datasets for training the model for table detection and table recognition tasks. We focus especially on the PubTables-1M dataset \cite{pubtable1m} as at the time of writing, it is the largest available non-synthetic research dataset in the domain. The datasets contain images of documents and files describing table structures. The PubTables-1M dataset contains XML files with coordinates of tables and table elements in the Pascal VOC format, while other datasets contain JSON files with cell coordinates and strings representing table structure in the HTML format.

To train the model, we have uniformized these datasets to the same format. For each document, we store a JSON file with the list of table elements names (table, rows, columns, headers), their coordinates, and the coordinates of spanning cells that should be included in those elements.

The PubTables-1M dataset consists of two separate datasets. One contains only labeled tables for the table detection task and another one contains only cropped-out tables with table structure labels. In that case, once the table is detected the model can concentrate on recognizing its structure. It was possible to map most, although not all, of the coordinates from the table recognition dataset onto the table detection dataset.

The authors of all these datasets have already split them into training, evaluation, and test sets. For large datasets, to reduce experiment runtime, we have limited the test set to at most 10000 documents. When comparing, we have observed consistently similar scores on the smaller and full sets.

The PubTables-1M dataset comes with word contents and coordinates. In our experiments for that dataset, we do not use an OCR engine by instead relying on these extracted coordinates to avoid the quality of the used OCR engine affecting results. For other datasets, we had to run an internally developed OCR engine first to extract word boxes and their contents. In a production setting, one might want to make sure that the same OCR engine is used during training and inference, as the results might be affected by OCR choices such as padding, orientation, etc.

We use labels for tables, rows, columns, and headers. We do not detect spanning cells separately as in \cite{pubtable1m}, but expand the dimensions of rows, columns, and headers by spanning cells. The PubTables-1M dataset contains also information about row headers that we do not consider. It is worth noting that the FinTabNet dataset does not have labels for headers, so during the training we have to be careful to mask the loss for that class.

For evaluation, we have also used the modern part (without historical images) of the ICDAR-2019 dataset \cite{icdar2019}.

\section{Experiments}
\label{sec:data-and-exp}

\subsection{Training details}
\label{sec:implementation}
We trained the model for the table detection and table recognition task. In this way, we could easily compare the performance with the models presented in \cite{pubtable1m}, at least on some of the classes.

We used the Adam optimizer with a learning rate of $10^{-4}$ for 100 epochs of 5000 steps each, batch size 8, and then decreased the learning rate to $10^{-5}$ in order to further tune the model for 100 more epochs.

\subsection{Model inputs and outputs}
As described in Section \ref{subsec:pos_enc_voc_emb}, we embed both bounding box coordinates and contents and sum the results together. We also add the encoding of image patches on top, whenever we use those. The input tensor to the transformer has then shape $(B=8, n_{words}=1000, d_{model}=256)$. For each task/class, the output of the model then has shape $(B, n_{words}, n_{words})$, which is also the same shape as the labels (ground truth adjacency matrices).

\subsection{Metric}
\label{sec:metric}

For validation purposes, we use the Dice score as a metric on the adjacency matrices. Similar to image segmentation problems, these matrices have a vast majority of 0 entries, which we therefore exclude from the count by using the metric:
$$
Dice = 2\frac{\#\{(i, j)\ |\ pred_{ij} = gt_{ij} = 1\}}{\#\{(i, j)\ |\ pred_{ij} = 1\} + \#\{(i, j)\ |\ gt_{ij} = 1\}}
$$

For testing purposes, we also include the standard average precision $AP[IoU=0.50]$ and $AP[IoU=0.50:0.95]$ and average recall scores as computed by the CocoEvaluator class distributed by torchvision. However, since our model can only predict bounding boxes around predicted words, prior to computing these metrics we first shrink ground truth cells, rows, etc. to the minimum bounding box surrounding words including in the original label boxes. Because of this reason, our metrics are not directly comparable to those in the literature. Finally, for the comparison to make sense, when evaluating table recognition we ignore spanning cells and only consider the 4 most common classes (cells, rows, columns, headers).


\subsection{Results}
\subsection{Considered results}
\label{sec:results}

We compare the performance of our model with the performance of the models in the paper \cite{pubtable1m} and present the numbers in Table \ref{tab:results_recognition} and Table \ref{tab:results_detection}. We also report the results of the same DETR model achieving state-of-the-art results in \cite{pubtable1m} retrained using the code in the GitHub repository provided by the authors, as well as the provided checkpoint, showing that achieving the same accuracy as reported in the paper is not always straightforward. In Table \ref{tab:results_different_datasets}, we report results on different datasets.

\subsection{Quantitative results}
\label{sec:quantitative_results}

We have tested several variants of the model, but overall results tended to be similar (more layers, larger hidden dimension, or larger output head dimension did not seem to significantly improve results), see section \ref{sec:ablation}. Including the image patches as inputs had a measurable impact on quality, at the price of a bigger and slower model.

\subsubsection{Table recognition}
We took the code from the Github of the paper \cite{pubtable1m} and retrained the DETR model ourselves with the same parameters specified in the author's Github. We have trained the DETR model for 20 epochs (1 week on a V100 GPU). In our model, we consider 4 classes (table, column, row, and header; our definition of spanning cell is not directly comparable with the one used in \cite{pubtable1m}), so we rerun the DETR model evaluation for these classes only. Our proposed model (using image patches) reached slightly worse results (AP 0.931) than the DETR model using the pre-trained checkpoint from GitHub (0.948), but better than the DETR model retrained by us with the provided code (0.888). Similar results are obtained for average recall (see Table \ref{tab:results_recognition}).

Excluding image patches from our model decreased scores by around 0.03. Better results without image patches can be obtained by skipping word normalization and building a dataset-specific vocabulary; however, we avoid doing that, because then it becomes difficult to generalize to other languages and other datasets, and we risk overfitting to the specific layout of the training dataset(s). Since image patches also carry information about table borders and bold text, it is still beneficial even with a dedicated vocabulary.

\subsubsection{Table detection}
For the table detection task, our model scored better than the DETR model. Our proposed model (using image patches) obtained 0.989 AP and 0.994 AR. This time, we have observed similar results for the DETR model trained by us and the DETR model from the Github of the paper \cite{pubtable1m}.

The task is much more straightforward than the table recognition task and all the models get very close to 1.0 average precision (see Table \ref{tab:results_detection}). Our model obtained better results in terms of average precision, as well as average recall.

\begin{table}
\centering
\caption{Results reported for the table structure recognition task. Comparison of our clustering model with the DETR model from \cite{pubtable1m}. 4 classes are taken into consideration as spanning cells and row headers are not present in other training datasets. Scores for the best model were highlighted in bold and for our model (second-best) scores were italicized.}
\label{tab:results_recognition}
\begin{tabular}{l|c@{\hskip 0.1in}c@{\hskip 0.1in}c@{\hskip 0.1in}|c@{\hskip 0.1in}c@{\hskip 0.1in}c@{\hskip 0.1in}|c@{\hskip 0.1in}c@{\hskip 0.1in}c}
\multirow{2}{*}{Model} & \multicolumn{3}{c}{4 classes} & \multicolumn{3}{c}{6 classes}\\
& AP & AP$_{50}$ & AR & AP & AP$_{50}$ & AR \\
\hline
Faster R-CNN \cite{pubtable1m} (paper)       &-&-&- & 0.722 & 0.815 & 0.762 \\
DETR \cite{pubtable1m} (paper)               &-&-&- & 0.912 & 0.971 & 0.942 \\
DETR \cite{pubtable1m} (Github)              &-&-&- & 0.902 & 0.970 & 0.935 \\
\hline
\multicolumn{7}{c}{Evaluation on the first 10000 files from the test set} \\
\hline
DETR (GitHub)                   & \textbf{0.948} & 0.989 & \textbf{0.966}
                                & 0.886 & 0.965 & 0.923 \\

DETR (trained by us for 1 week)  & 0.888 & 0.983 & 0.922
                                & 0.818 & 0.954 & 0.873 \\
\hline
Clustering model w/ images      & \textit{0.931} & 0.972 & \textit{0.952}
                                & - & - & - \\
Clustering model w/out images   & 0.891 & 0.941 & 0.923 
                                & - & - & -
\end{tabular}
\end{table}

\begin{table}
\centering
\caption{Results reported for the table detection task. Comparison of our clustering model with the DETR model from \cite{pubtable1m}.}
\label{tab:results_detection}
\begin{tabular}{l|c@{\hskip 0.1in}c@{\hskip 0.1in}c}
Model & AP & AP$_{50}$ & AR \\
\hline
Faster R-CNN \cite{pubtable1m} (paper) & 0.825 & 0.985 & 0.866 \\
DETR \cite{pubtable1m} (paper)         & 0.966 & 0.995 & 0.981 \\
DETR \cite{pubtable1m} (Github)        & 0.970 & 0.995 & 0.985 \\
\hline
\multicolumn{4}{c}{Evaluation on the first 10000 files from the test set} \\
\hline
DETR (GitHub)                   & 0.958 & 0.995 & 0.977 \\
DETR (trained by us for 1 week) & 0.949 & 0.995 & 0.974 \\
\hline
Clustering model w/ images      & \textbf{0.989} & 0.990 & \textbf{0.994} \\
Clustering model w/out images   & 0.982 & 0.990 & 0.988
\end{tabular}
\end{table}

\begin{table}
\centering
\caption{Results reported for the table detection and recognition task for our clustering model trained on different datasets at once. The evaluation was performed on the first 10000 files, or all files if fewer were available, from the test set}
\label{tab:results_different_datasets}
\begin{tabular}{l|c@{\hskip 0.1in}c@{\hskip 0.1in}c}
Test Dataset & AP & AP$_{50}$ & AR \\
\hline
PubTables-1M \cite{pubtable1m}                     & 0.924 & 0.951 & 0.945  \\
FinTabNet \cite{global_table_extractor}            & 0.792 & 0.859 & 0.875 \\
PubTabNet \cite{image_based_table_recognition}     & 0.921 & 0.953 & 0.947 \\
ICDAR-2019 \cite{icdar2019}                        & 0.746 & 0.835 & 0.820
\end{tabular}
\end{table}

\pagebreak
\subsection{Ablation study}
\label{sec:ablation}

We have trained several models with different configurations. We considered the influence of the following model properties:
\begin{itemize}
    \item Transformer number of layers
    \item Dimension of the last fully connected output layer
    \item Embedding dimension of inputs
    \item Using or skipping image patches
\end{itemize}

We found out that the models trained in different configurations listed in Table \ref{tab:results_recognition_ablation_study} perform comparably. On the architecture side, we only observed a small decrease in quality when using a very small model with a two-layer transformer with $d_{model}$=128. As already noted in section \ref{sec:quantitative_results}, the presence of image patches has a more significant effect.

We have also experimented with a joint model trained on table detection and table recognition datasets at the same time. The performance of such a model was comparable to other models. That results in a lower runtime and lower memory consumption, as we do not require separate models for table detection and table recognition tasks.

\begin{table}
\centering
\caption{Results reported for the table structure recognition task for different model configurations. Layer \# is the number of transformer encoder layers, $C_{out}$ is the output dimension of the last layer (i.e. both $Q$ and $K$ embed in dimension $C_{out} / 2$), dff and $d_{model}$ are transformer parameters.}

\label{tab:results_recognition_ablation_study}
\begin{tabular}{c|cccccc|c@{\hskip 0.1in}c@{\hskip 0.1in}c|}
\multirow{2}{*}{Image patches} & \multirow{2}{*}{Layer \#} & \multirow{2}{*}{dff} & \multirow{2}{*}{$C_{out}$} & \multirow{2}{*}{$d_{model}$} & \multirow{2}{*}{\shortstack{Total \\ Param \#}} & \multirow{2}{*}{\shortstack{Param \# exc. \\ embeddings}} & \multicolumn{3}{c}{4 Classes} \\
& & & & & & & AP & AP$_{50}$ & AR \\
\hline
Yes & 4 & 1024 & 300 &  256 & 15M & 5M & \textbf{0.931} & \textbf{0.972} & \textbf{0.952} \\
Yes & 4 & 1024 & 1000 & 256 & 19M & 10M & 0.926 & 0.971 & 0.949 \\
Yes & 4 & 1024 & 300 & 512 & 32M  & 12M & \textbf{0.932} & \textbf{0.975} & \textbf{0.952} \\
Yes & 8 & 1024 & 300 & 256 & 18M  & 8M & 0.929 & 0.969 & 0.949 \\
Yes & 2 & 1024 & 300 & 128 & 7M   & 2M & 0.912 & 0.959 & 0.937 \\
No & 4 & 1024 & 300 &  256 & 14M & 4M & 0.891 & 0.941 & 0.923 \\
\end{tabular}

\end{table}

\pagebreak

\subsection{Examples}
\begin{figure}
\begin{subfigure}[t]{0.48\textwidth}
    \includegraphics[width=\textwidth]{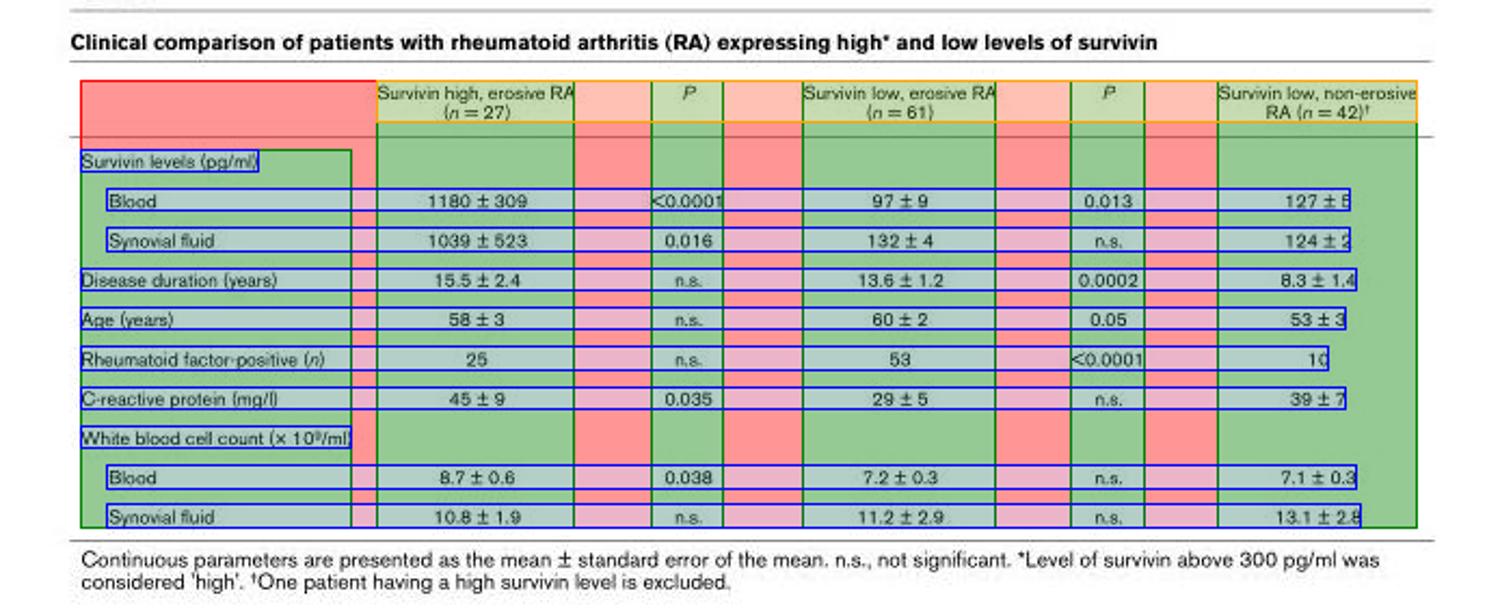}
\label{subfig:a}

\end{subfigure}\hspace{\fill} 
\begin{subfigure}[t]{0.48\textwidth}
    \includegraphics[width=\linewidth]{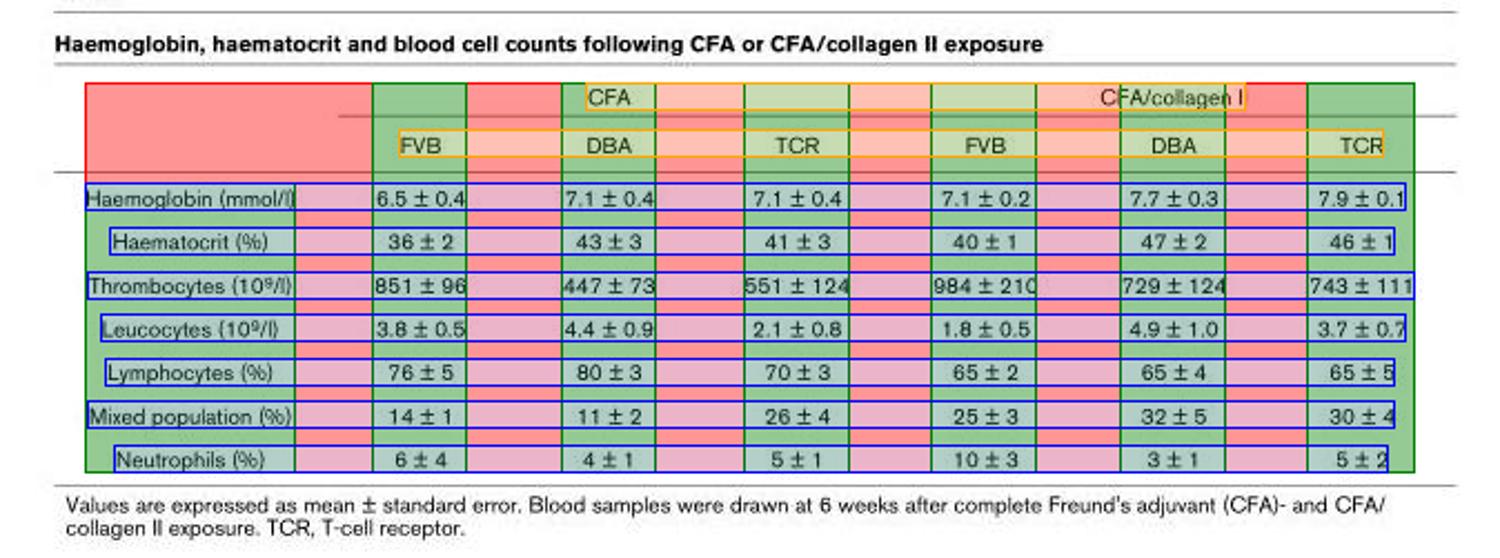}
\label{subfig:b}
\end{subfigure}

\begin{subfigure}[t]{0.35\textwidth}
    \includegraphics[width=\textwidth]{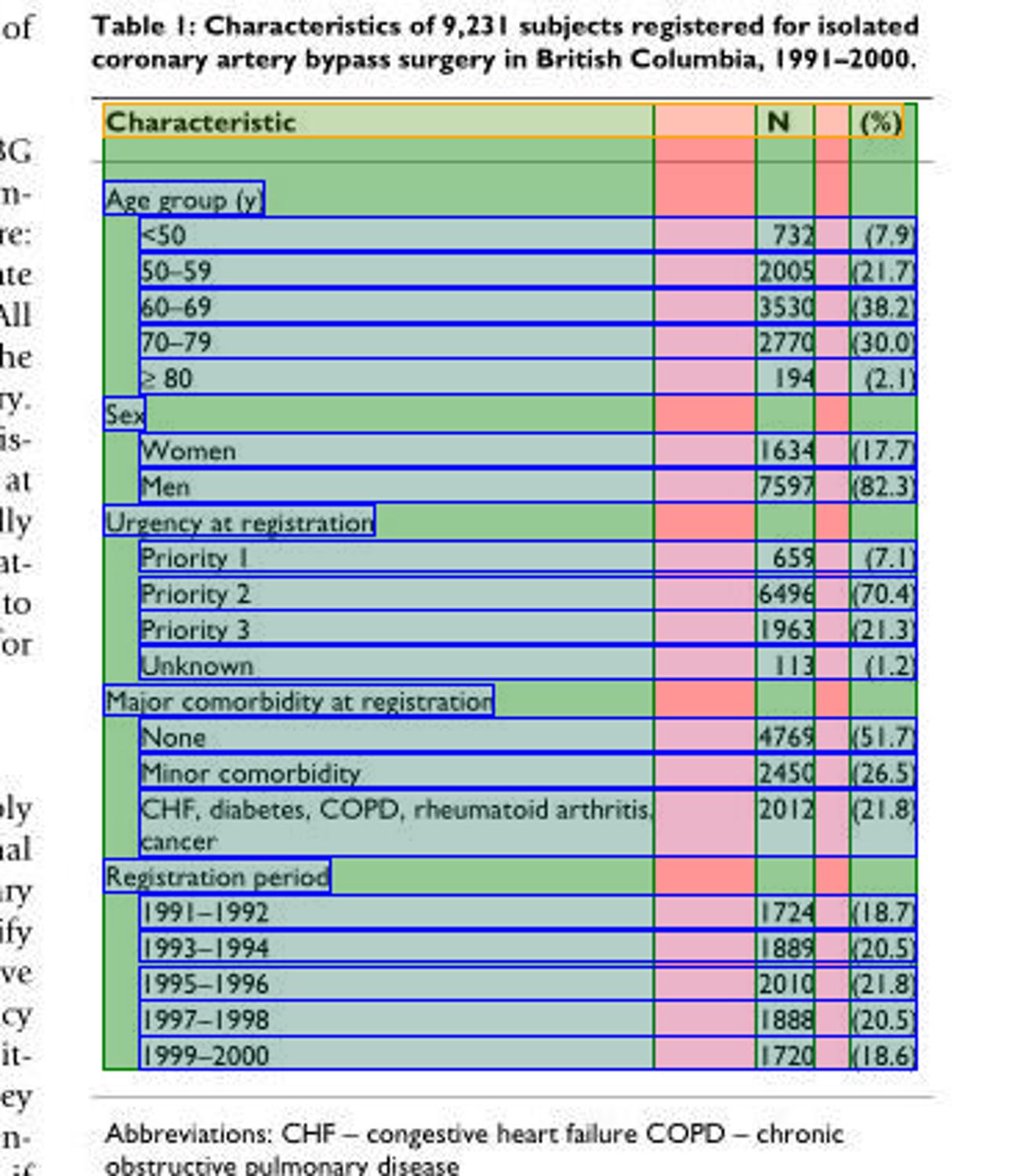}
\label{subfig:c}

\end{subfigure}\hspace{\fill} 
\begin{subfigure}[t]{0.45\textwidth}
    \includegraphics[width=\linewidth]{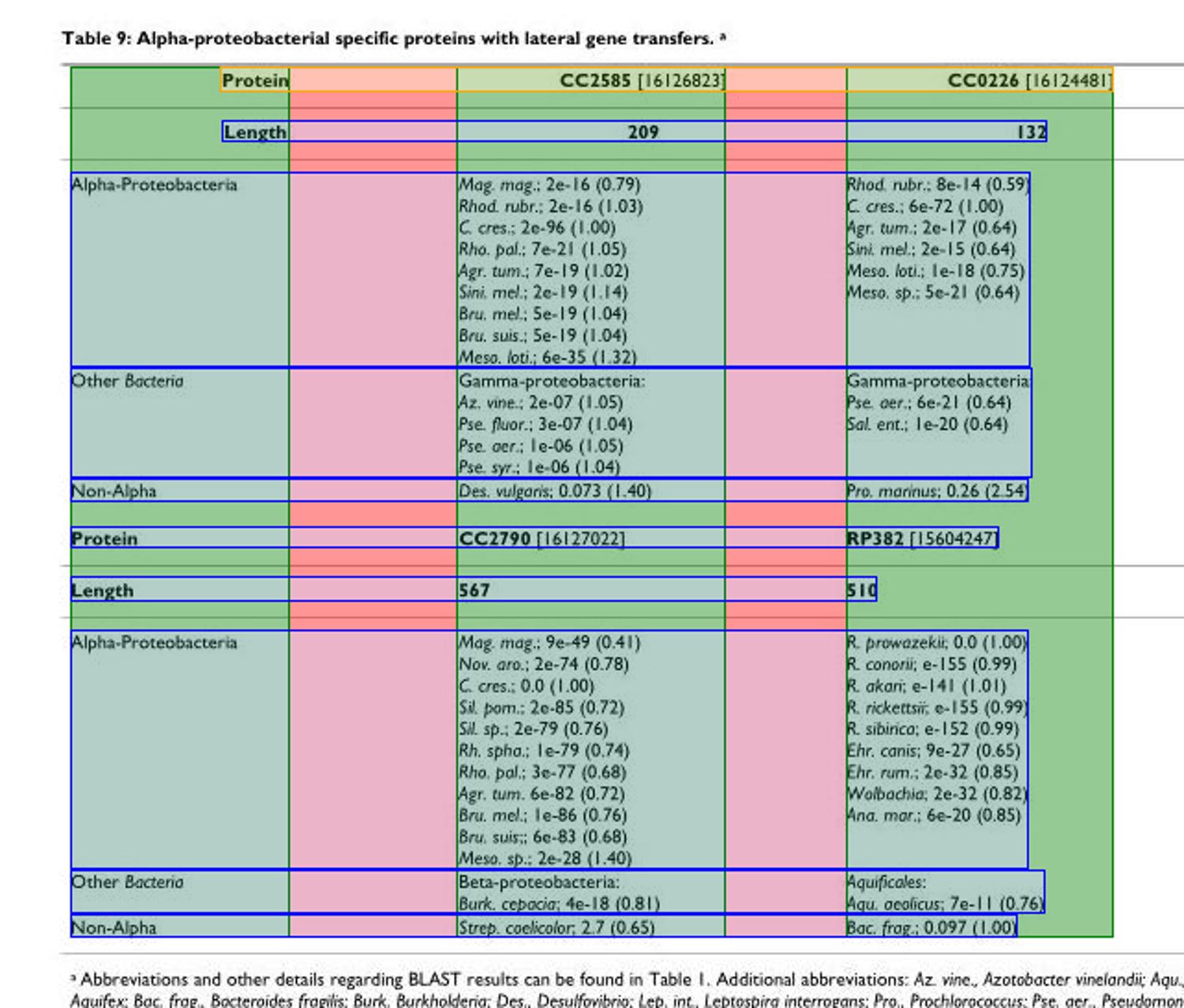}
\label{subfig:d}
\end{subfigure}

\caption{Examples from the test set showcasing some correct predictions.}
\label{fig:correct_predictions}

\begin{subfigure}[t]{0.48\textwidth}
    \includegraphics[width=\textwidth]{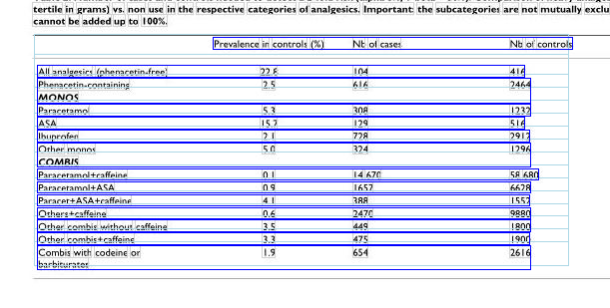}
\caption{Two rows merged together}
\label{subfig:a}

\end{subfigure}\hspace{\fill} 
\begin{subfigure}[t]{0.48\textwidth}
    \includegraphics[width=\linewidth]{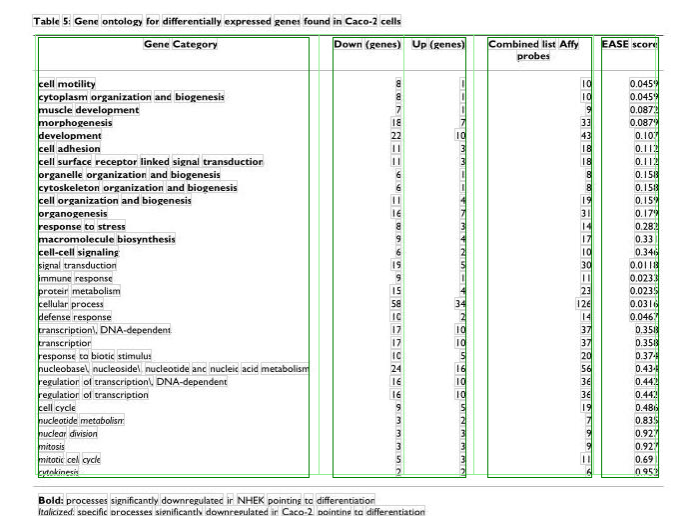}
\caption{Two columns merged together}
\label{subfig:b}
\end{subfigure}

\caption{Examples showcasing common errors, taken from the test set}
\label{fig:errors}
\end{figure}

\begin{figure}
\begin{subfigure}[t]{0.50\textwidth}
    \includegraphics[width=\textwidth]{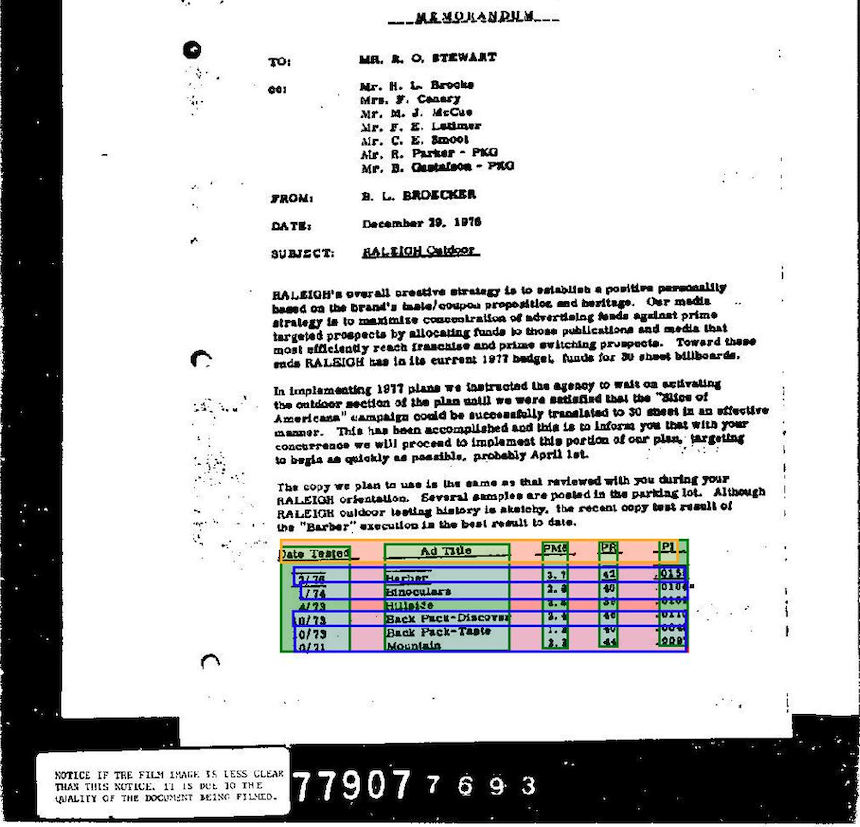}
\label{subfig:challenging1}

\end{subfigure}\hspace{\fill} 
\begin{subfigure}[t]{0.50\textwidth}
    \includegraphics[width=\linewidth]{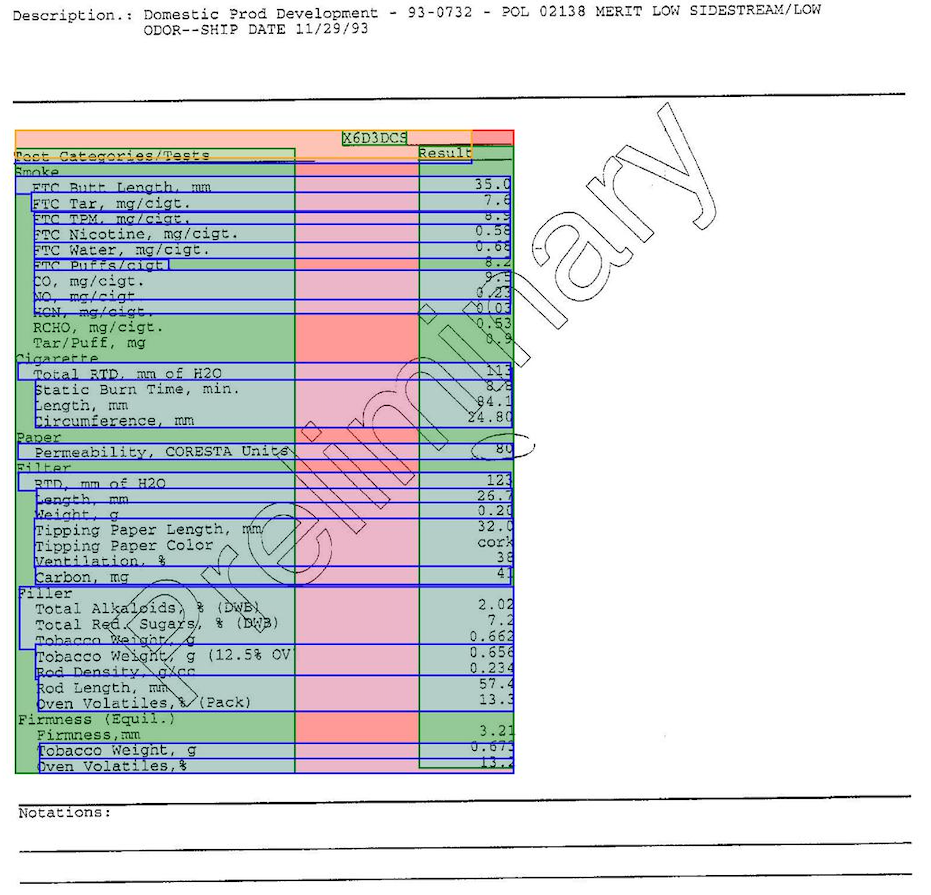}
\label{subfig:challenging2}
\end{subfigure}

\caption{Examples showcasing table detection and recognition for challenging documents from the Tobacco dataset \cite{tobacco}. Several rows are wrongly skipped or joined, but the overall performance is satisfying, considering that the model was only trained on images from ``clean'' modern documents.   t}
\label{fig:challenging}
\end{figure}

In Figure \ref{fig:correct_predictions}, we visualize some of the correct predictions. Some of the most common mistakes we could observe are merging two rows or columns together (see Figure \ref{subfig:a} and Figure \ref{subfig:b}). Challenging, out-of-distribution examples are visualized in Figure \ref{fig:challenging}.

\clearpage
\vfill
\clearpage

\section{Conclusion}
We have presented the new method of table detection and table structure recognition. A core novelty of our model is the technique of supervised clustering with transformer networks and its application to any object detection task for documents. Although our method was applied to table detection it could cluster words into other categories given the labeled dataset. Moreover, it could be used in other domains given well-defined input elements such as word boxes in the considered problem.

We show empirically that our model is on par with DETR from \cite{pubtable1m}. Our model is much smaller and relies on the OCR output instead of only images. Since OCR engines are already optimized to detect text in noisy documents, we do not need to worry about image data augmentation. Moreover, we can handle skewed/rotated documents as we can calculate the skewed/rotated bounding box around the words in the cluster, and our output format (the adjacency matrix) is completely agnostic to document rotation.

\bibliographystyle{splncs04}
\bibliography{references}

\begin{thebibliography}{10}
\providecommand{\url}[1]{\texttt{#1}}
\providecommand{\urlprefix}{URL }
\providecommand{\doi}[1]{https://doi.org/#1}

\bibitem{transformer_detection}
Carion, N., Massa, F., Synnaeve, G., Usunier, N., Kirillov, A., Zagoruyko, S.:
  End-to-end object detection with transformers. In: European conference on
  computer vision. p. 213–229 (2020)

\bibitem{detr}
Carion, N., Massa, F., Synnaeve, G., Usunier, N., Kirillov, A., Zagoruyko, S.:
  End-to-end object detection with transformers. In: European Conference on
  Computer Vision. p. 213–229 (2020)

\bibitem{tree_bases}
Cesarini, F., Marinai, S., L.~Sarti, G.S.: Trainable table location in document
  images. In: Object recognition supported by user interaction for service
  robots. p. 236–240 (2002)

\bibitem{complicated_table_structure_recognition}
Chi, Z., Huang, H., Xu, H.D., Yu, H., Yin, W., Mao, X.L.: Complicated table
  structure recognition. arXiv preprint arXiv:1908.04729  (2019)

\bibitem{icdar2019}
Gao, L., Y.~Huang, H.D., Meunier, J.L., Yan, Q., Fang, Y., Kleber, F., Lang,
  E.: Icdar 2019 competition on table detection and recognition (ctdar). In:
  2019 International Conference on Document Analysis and Recognition (ICDAR).
  p. 1510–1515 (2019)

\bibitem{faster_rcnn_based_model}
Gilani, A., Qasim, S.R., Malik, I., Shafait, F.: Table detection using deep
  learning. In: 2017 14th IAPR international conference on Document Analysis
  and Recognition (ICDAR). p. 771–776 (2017)

\bibitem{yolo}
Huang, Y., Yan, Q., Li, Y., Chen, Y., Wang, X., Gao, L., Tang, Z.: A yolo-based
  table detection method. In: 2019 International Conference on Document
  Analysis and Recognition (ICDAR). p. 813–818 (2019)

\bibitem{unsupervised_clustering}
Kieninger, T., Dengel, A.: The t-recs table recognition and analysis system.
  In: International Workshop on Document Analysis Systems. p. 255–270 (1998)

\bibitem{tobacco}
Lewis, D., Agam, G., Argamon, S., Frieder, O., Grossman, D., Heard, J.:
  Building a test collection for complex document information processing. In:
  Proceedings of the 29th annual international ACM SIGIR conference on Research
  and development in information retrieval. pp. 665--666 (2006)

\bibitem{tablebank}
Li, M., Cui, L., Huang, S., Wei, F., Zhou, M., Li, Z.: Tablebank: Table
  benchmark for image-based table detection and recognition. In: 12th Language
  Resources and Evaluation Conference. p. 1918–1925 (2020)

\bibitem{heuristic_spacing}
M.~A.~Jahan, R.G.R.: Locating tables in scanned documents for reconstructing
  and republishing. In: 7th International Conference on Information and
  Automation for Sustainability. pp.~1--6 (2014)

\bibitem{synthtabnet}
Nassar, A., Livathinos, N., M.~Lysak, P.S.: Tableformer: Table structure
  understanding with transformers. In: IEEE/CVF Conference on Computer Vision
  and Pattern Recognition. p. 4614–4623 (2022)

\bibitem{tablenet}
Paliwal, S.S., Vishwanath, D., Rahul, R., Sharma, M., Vig, L.: Tablenet: Deep
  learning model for end to-end table detection and tabular data extraction
  from scanned document images. In: 2019 International Conference on Document
  Analysis and Recognition (ICDAR). p. 128–133 (2019)

\bibitem{cascadetabnet}
Prasad, D., Gadpal, A., Kapadni, K., Visave, M., Sultanpure., K.:
  Cascadetabnet: An approach for end to end table detection and structure
  recognition from image-based documents. In: IEEE/CVF Conference on Computer
  Vision and Pattern Recognition Workshops. p. 572–573 (2020)

\bibitem{rethinking_table_recognition_using_graph_neural_networks}
Qasim, S.R., Mahmood, H., Shafait, F.: Rethinking table recognition using graph
  neural networks. In: 2019 International Conference on Document Analysis and
  Recognition (ICDAR). p. 142–147 (2019)

\bibitem{faster_rcnn}
Ren, S., He, K., Girshick, R., Sun, J.: Faster r-cnn: Towards real-time object
  detection with region proposal networks. arXiv preprint arXiv:1506.01497
  (2015)

\bibitem{deepdesrt}
Schreiber, S., Agne, S., Wolf, I., Dengel, A., Ahmed, S.: Deepdesrt: Deep
  learning for detection and structure recognition of tables in document
  images. In: 2017 14th IAPR international conference on document analysis and
  recognition (ICDAR). p. 1162–1167 (2017)

\bibitem{pubtable1m}
Smock, B., Pesala, R., Abraham, R.: Pubtables-1m: Towards comprehensive table
  extraction from unstructured documents. In: IEEE/CVF Conference on Computer
  Vision and Pattern Recognition (CVPR). pp. 4634--4642 (2022)

\bibitem{decision_tree_svm}
Wang, Y., Hu, J.: A machine learning based approach for table detection on the
  web. In: 11th international conference on World Wide Web. p. 242–250 (2002)

\bibitem{heuristic_distance}
Wangt, Y., Phillipst, I.T., Haralick, R.: Automatic table ground truth
  generation and a background-analysis-based table structure extraction method.
  In: Sixth International Conference on Document Analysis and Recognition. p.
  528–532 (2001)

\bibitem{xu2019layoutlm}
Xu, Y., Li, M., Cui, L., Huang, S., Wei, F., Zhou, M.: Layoutlm: Pre-training
  of text and layout for document image understanding (2019)

\bibitem{global_table_extractor}
Zheng, X., Burdick, D., Popa, L., Zhong, X., Wang, N.X.R.: Global table
  extractor (gte): A framework for joint table identification and cell
  structure recognition using visual context. In: IEEE/CVF Winter Conference on
  Applications of Computer Vision. p. 697–706 (2021)

\bibitem{image_based_table_recognition}
Zhong, X., ShafieiBavani, E., Yepes, A.J.: Image-based table recognition: data,
  model, and evaluation. arXiv preprint arXiv:1911.10683  (2019)

\end{thebibliography}
\end{document}